\documentclass[akbc,twoside,11pt,lettersize]{article}
\usepackage{akbc}
\usepackage{latexsym}
\usepackage{todonotes}
\usepackage{url}
\usepackage{amsmath}
\usepackage{bbm}
\usepackage{framed}
\usepackage{amssymb}
\usepackage{xcolor}
\usepackage{cleveref}
\usepackage{multirow}

\newcommand{\ourapproach}{{\sc ours}} 

\newcommand{\conll}{CoNLL}
\newcommand{\tackbp}{TAC-KBP 2010}

\newcommand{\blank}{\textsc{[mask]}}
\newcommand{\ctx}{\mathbf{x}}
\newcommand{\spn}{\mathbf{s}}
\newcommand{\lbl}{\mathbf{l}}
\newcommand{\lblset}{\overline{\mathbf{l}}}
\newcommand{\spanstart}{s_{start}}
\newcommand{\spanend}{s_{end}}
\newcommand{\candidates}{\mathcal{C}_{\lblset}}

\makeatletter
\newcommand{\printfnsymbol}[1]{%
  \textsuperscript{\@fnsymbol{#1}}%
}
\makeatother

\akbcheading
\ShortHeadings{Empirical Evaluation of Pretraining Strategies for Supervised Entity Linking}{Févry, FitzGerald, Baldini-Soares, \& Kwiatkowski}
\finalcopy 
\begin{document}

\title{Empirical Evaluation of Pretraining Strategies for Supervised Entity Linking}

\author{\name Thibault F{\'e}vry\thanks{Denotes equal contribution.}~\thanks{Work conducted during Google AI Residency.} \email tfevry@google.com \\
       \name Nicholas FitzGerald\printfnsymbol{1} \email nfitz@google.com \\
       \name Livio Baldini Soares \email liviobs@google.com \\
       \name Tom Kwiatkowski \email tomkwiat@google.com \\
       \addr Google Research, NYC}

\maketitle

\begin{abstract}
In this work, we present an entity linking model which combines a Transformer architecture with large scale pretraining from Wikipedia links. Our model achieves the state-of-the-art on two commonly used entity linking datasets: 96.7\% on CoNLL and 94.9\% on TAC-KBP. We present detailed analyses to understand what design choices are important for entity linking, including choices of negative entity candidates, Transformer architecture, and input perturbations. Lastly, we present promising results on more challenging settings such as end-to-end entity linking and entity linking without in-domain training data. 
\end{abstract}

\section{Introduction}\label{sec:intro}

Traditionally, entity linking approaches have relied on knowledge bases, complicated modelling and task-specific hand-engineered features to achieve high performance.
More recently, \citealt{broscheit}, \citealt{ling2020learning} and~\citealt{zero_shot_el} show that using large-scale pretrained language models like BERT~\cite{devlin2018bert}, pretraining on Wikipedia entity links, and fine-tuning on a specific entity linking corpus leads to state-of-the-art performance without relying on such features.
However, \citealt{ling2020learning} focused mainly on constructing general-purpose entity representations,  \citealt{zero_shot_el} on building strong zero-shot entity linking systems, and~\citealt{broscheit} on end-to-end linking, so that the limits of pretraining for entity disambiguation have not been fully explored. 

In this paper we present a thorough study of pretraining strategies for supervised entity linking. We establish new upper bounds for performance on the widely studied \conll~and \tackbp~entity linking tasks. We also show that our pretraining approach yields a very competitive entity linking system without any further domain specific tuning.
We present a detailed analysis of significant design choices including the choice of negative candidates used during training, and the document context encoded for each mention. We find that the optimal choice of negative candidates is dependent on whether or not the final linking system has access to an alias table.
For a system that will use an alias table at inference, it is helpful to pretrain the models with lexically similar candidates. 
However, when no alias table is used for the downstream task, ensuring candidates are random improves the model ability to distinguish the right entity among all possible entities.

In our studies we found two surprising results: (1) it is possible to achieve optimal entity linking results with a four layer transformer, which is one third of the size of BERT-base, and (2) given the abundance of supervision from Wikipedia links, we did not get any gains in performance from training with an auxiliary language modeling loss. However, we did find that the input perturbations introduced by \cite{devlin2018bert}~themselves increase the quality of our pretraining approach and we present an analysis of how these perturbations add robustness to the model.

Finally, to demonstrate the generality of our model, we present results on the end-to-end entity linking task in which both mention location and identity are predicted. On this task, our model outperforms all but the tailored methods introduced by \citealt{Kolitsas2018-dn, broscheit}. We argue that our model is more practical, and easier to integrate, than heavily engineered existing approaches, and we believe that downstream tasks such as information extraction and question answering can benefit from this robust standalone end-to-end entity linker (e.g.~\cite{fevry2020eae}).

\section{Related Work}\label{sec:related_work}

Early entity linking systems \cite{bunescu2006using, mihalcea2007wikify} focused on matching the context of the mention with that of the entity page. In addition to context features, systems have relied on $\mathbb{P}(e|m)$, the prior probability that mention $m$ refers to entity $e$, computed from Wikipedia mention counts. The set of entities in a document should be globally coherent, and several approaches have introduced sophisticated global disambiguation methods \cite{globerson2016collective, cheng2013relational, sil2013re} that consider all mentions in a document to make predictions. In contrast, we do not model document-level disambiguation explicitly. However, our long context windows contain several mentions which should allow such disambiguation. Other approaches have also sought to incorporate types and knowledge base information in their modelling, such as \citealt{radhakrishnan,raiman2018deeptype}.

We pretrain our model by learning distributed representation of entities directly from Wikipedia text, similarly to \citealt{yamada2016joint,Yamada2017-dv,gillick2019learning, ling2020learning}. Unlike \citealt{zero_shot_el, gupta2017entity}, our embeddings are learned directly rather than generated through entity descriptions. In contrast with \citealt{yamada2016joint, Yamada2017-dv}, we do not use additional features, such as prior probabilities or string match features. Our method is therefore most similar to \citealt{ling2020learning} and \citealt{broscheit} who also use a transformer. We differ from \citealt{ling2020learning} by simultaneously considering all mentions and entities in a context and, in contrast to both, only use a four layer, randomly-initialized transformer instead of twelve layers initialized from large scale language modelling pretraining. 

In addition, we experiment with end-to-end entity linking \cite{sil2013re,luo2015joint,Kolitsas2018-dn}, where instead of predicting the entity for gold spans, a system must both predict the span and its label. A closely related task is multilingual entity linking. Approaches have used multilingual embeddings to link text in several languages \cite{sil2018neural,tsai2016cross}. Zero-shot entity linking \cite{logeswaran2019zero,zero_shot_el} is another relevant task. In that setting, entities predicted at test time are not seen in training. Instead, the system relies on the entity name and description.

\section{Model}\label{sec:model}

    \subsection{Task Definition}

    Let $\mathcal{E} = \{e_1 \dots e_N\}$ be a predefined set of entities, and let $\mathcal{V} = \{\blank, w_1 \dots w_M\}$ be a vocabulary of words.
    A \emph{context} $\ctx = [x_0 \dots  x_t]$ is a sequence of words $x_i \in \mathcal{V}$.
    A \emph{span} $\spn = (\spanstart, \spanend)$, is a tuple with $0 \leq \spanstart, \spanend < t$ which defines a contiguous sequence of tokens in a 
    given context.
    A \emph{mention label} $\lbl = (s_k, e_k)$ consists of a span $s_i$ and an entity label, $e_i \in \mathcal{E} \cup \varnothing$.
    We use $\lblset$ to denote a set of such mention labels.
    The \texttt{NULL}-symbol $\varnothing$ indicates a span that is labeled as a mention, but without an entity linking label.
    
    Our training data, $\mathcal{D} = \{(\ctx_0, \lblset_0) \dots (\ctx_N, \lblset_N)\}$, is a corpus of contexts, each paired with a \emph{set} of mention labels, one for each mention in the context.
    Given an input context $\ctx_i$, our goal is to predict the set of entity mentions $\lblset_i$.
    In \emph{Entity Disambiguation}, we are given the set of spans, and predict the entity linked by each span.
    In \emph{End-to-End Entity Linking}, we must predict both the set of mention spans, and their linked entities.

    \subsection{Contextual Language Representation}

    Our model is built using the now-standard Transformer-based architecture~\citep{vaswani2017attention}. 
    The model computes a matrix representation $\hat{\mathbf{H}} \in \mathcal{R}^{t \times d}$ of a text sequence through successive application of a Transformer block to the output of the previous layer:
    
    \begin{align*}
        \mathbf{H}_{i} &= \mathtt{TransformerBlock}(\mathbf{H}_{i-1}) \\
        &= \mathtt{MLP}(\mathtt{MultiHeadAttention}(\mathbf{H}_{i-1}, \mathbf{H}_{i-1}, \mathbf{H}_{i-1}))
    \end{align*}
    
    $\mathbf{H}_0$ is a sequence of context-independent token embeddings and $\hat{\mathbf{H}} = \mathbf{H}_n$, where $n$ is the number of Transformer layers.
    
    \subsection{Entity Disambiguation}
    
    Each entity $e \in \mathcal{E}$ is mapped directly onto a dedicated vector in $\mathbb{R}^d$ via a $\mathcal{|\mathcal{E}|} \times d$ dimensional embedding matrix.
    In our experiments, we have a distinct embedding for every concept that has an English Wikipedia page, resulting in approximately 5.7m entity embeddings.
    
    In order to perform entity linking for a particular span with word-piece token indices $(i, j)$, we (following~\citealt{lee2016learning}) first obtain a representation of the span by concatenating the representation at the span start and end, and pass this through a multi-layer perceptron which projects the span representation into the same space as the entity embeddings.
    
    \begin{equation}
        \hat{s}_{\spn_i} = \texttt{MLP}([\mathbf{H}_{n, \spanstart}, \mathbf{H}_{n, \spanend}])
    \end{equation}
    
    Our model scores each span-entity pair by taking the dot-product between the projected span representation and the embedding of $\mathbf{e_c}$. Thus, the conditional probability that the span $\spn_i$ refers to entity $e_c$ is defined as:
    
    \begin{equation}
     \mathbb{P}(e_c | \spn_i) = \frac{\exp(\hat{s}_{\spn_i} \cdot e_c)}{\sum_{\mathbf{c'} \in \mathcal{E}}\exp(\hat{s}_{\spn_i} \cdot e_{c'})}    
    \end{equation}
    
    In practice, this is expensive to compute for large $|\mathcal{E}|$. Therefore, for every $\lblset$ we select a set $\candidates$ of $k$ candidates, which contains the entity labels for all $\lbl \in \lblset$ as well as a set of negative candidates. We do not have an entity linking loss on mentions that do not have a label. Therefore, our per-example entity linking loss is:
    
    \begin{equation}
        l_{linking}(\lblset) = \sum_{\lbl_i \in \lblset} \frac{\exp(\hat{s}_{\spn_i} \cdot e_{i}) \mathbbm{1}_{e_i != \varnothing}}  
        {\sum_{c \in \candidates} \exp(\hat{s}_{\spn_i} \cdot e_{c})}
    \end{equation}
    
    We will discuss further how we select $\candidates$ in Section~\ref{sec:wiki_pretraining}.
    
    \subsection{Mention Detection}
    
    For many entity linking tasks, the target spans are provided.
    In order to be able to do end-to-end entity linking, we additionally train our model to predict mentions, independently of entity linking.
    One way to do this would be to score every possible span-entity pair, and either use a score threshold to filter spans where no entity link achieves a sufficiently high score, or to additionally score a special \texttt{NULL}-link embedding.
    However, enumerating all spans for the long contexts we use in our model would be prohibitively expensive.
    We take the approach of encoding mentions as a BIO sequence, and train an MLP on the context representation to predict this sequence with a standard cross-entropy loss. Our final loss sums the mention detection loss and the linking loss.

\section{Experimental Setup}\label{sec:experimental_setup}

\subsection{Wikipedia Pretraining}\label{sec:wiki_pretraining}

We build a training corpus of contexts paired with entity mention labels from the 2019-04-14 dump of English Wikipedia.
We first divide each article into chunks of 1000 unicode characters, resulting in a corpus of over 17.5 million contexts with over 17 million entity mentions covering over 5.7 million entities. These are processed with the BERT tokenizer, limited to 256 word-piece tokens.
In addition to the Wikipedia links, we annotate each sentence with unlinked mention spans using a state-of-the-art named entity recognizer.
These are used as additional signal for our mention detection component.

\paragraph{Entity Candidates Selection}

Training the model with a full softmax over all 5.7 million entities for every mention is computationally expensive. A common solution is to use a noise contrastive loss \cite{gutmann2012noise, mnih2013learning} and sample candidates according to their relative frequency, as in \citet{ling2020learning}. In this work, we experiment with other approaches to candidate generation that might provide better negatives in training. In addition to negatives selected uniformly at random from the entire entity vocabulary, we define two types of hard negatives:

\begin{enumerate}
    \item \textbf{Page candidates}, which is the set of all entities linked to in the article from which the given context was taken. This is meant to capture semantically related concepts.
    \item \textbf{Phrase table candidates}, the set of lexically related entities for each mention candidate, obtained from the Phrase Table provided by SLING~\citep{ringgaard2017sling}.
\end{enumerate}

Throughout the paper, we will use $|\candidates| = 768$. In our base setup, we use up to 256 page candidates, and 384 phrase table candidates, equally divided between each mention in the example. Any remaining room in the set of 768 is filled with random candidates sampled uniformly from the entity vocabulary (meaning a minimum of 128 random candidates per example). We will study the impact of different candidate selection methods in Section~\ref{sec:candidate_selection}. In addition to those candidates, for every example in a batch, we use the candidates of other examples as additional negatives. 

\paragraph{Input Noising} We also add noise to the input data. We apply the same noise function as is used in \citealt{devlin2018bert}: 15\% of the tokens are chosen to be modified. 80\% of those tokens are changed to the \blank~token, 10\% are changed to a random token and 10\% are left unmodified. 

\paragraph{Pretraining hyperparameters} We use \textsc{Adam} \cite{kingma2014adam} with a learning rate of 1e-4 to optimize our model. We use a linear warmup schedule for the first 10\% of training, decay the learning rate afterwards and use gradient clipping with a norm of 1.0. We train from scratch for up to a million steps and use a large batch size of 8192 for pretraining.  We follow BERT~\citep{devlin2018bert} base for many of our model parameters, though we do not use large-scale language-modeling pretraining and only use four layers, as we did not find more to layers to further improve performance. We use the same word-piece vocabulary as the lowercase version of BERT. We use entity embeddings of size 256 unless mentioned otherwise. We weight both the entity disambiguation loss and the mention detection loss to 1. We use a context window of 256 tokens. 

\subsection{Entity Linking Datasets}

We evaluate our model on two popular entity linking benchmarks: AIDA CoNLL YAGO~\cite{Hoffart2011-ip} and TAC-KBP 2010~\cite{ji2010overview}. The first is comprehensively annotated with approximately 34,000 mentions on 1,393 newswire document on the full Wikipedia vocabulary, while the second is sparsely annotated for target entities only on a smaller entity vocabulary.

\subsubsection{AIDA CoNLL-YAGO Dataset}

\paragraph{Textual Context} Most CoNLL documents do not fit in our limit of 256 tokens. Therefore, we split the document into ``sentences'' at each newline in the document. We experiment with three methods to add document context to these sentences: (i) taking the sentence as-is, (ii) adding the title of the document to the sentence, (iii) adding the title of the document as well as the first two sentences to the sentence. Throughout our experiments we will use (iii), though we show the impact of this choice in Section~\ref{sec:conll_context_selection}.

\paragraph{Entity Candidates Selection}
Our candidates for CoNLL come from alias tables - resources which provide a list of possible strings for a given entity.
A key challenge with evaluating entity linking systems on the CoNLL dataset is inconsistent use of alias tables.
~\citealt{globerson2016collective} describe the difficulty of resolving older resources due to changes in Wikipedia links and unicode, and provide statistics for two commonly-used alias tables: The YAGO extended ``means'' mapping of ~\citealt{Hoffart2011-ip}, and the ``PPRforNED'' mapping of ~\citealt{pershina2015personalized}.

We find that through careful resolution of unicode and Wikipedia redirects, we achieve a slightly higher conversion rate than reported by~\citealt{globerson2016collective} (statistics provided in Table~\ref{tab:at-stats}).
This leads to a higher gold recall, but also a larger number of candidate for each mention, meaning our system must distinguish between more candidates.
We report results using both alias tables.

\paragraph{Finetuning} We finetune our model -- including the entity embeddings -- on the CoNLL training set,
using the alias table candidates for each mention. We used a batch size of 256, a learning rate of 1e-6, and train for 2000 steps.

\begin{figure}
    \centering
    \begin{tabular}{|c|c|c|c|}
         \hline
        Alias Table & Conversion & Gold recall & Avg. ambig. \\
        \hline
         \multirow{2}{*}{~\citealt{Hoffart2011-ip}} & ~\citealt{globerson2016collective} & 96.19 & 65.9 \\
         & Ours & 99.33 & 67.4 \\
         \hline
         \multirow{2}{*}{~\citealt{pershina2015personalized}} & ~\citealt{globerson2016collective} & 99.84 & 12.6 \\
         & Ours & 99.75 & 13.8 \\
         \hline
    \end{tabular}
    \caption{Statistics for alias table conversions, computed on the CoNLL test split. Gold recall is the percentage of mentions for which the gold entity is included in the candidate set. Average ambiguity is the total number of candidates divided by the number of mentions.}
    \label{tab:at-stats}
\end{figure}

\subsubsection{\textsc{TAC-KBP} 2010 dataset}

TAC-KBP 2010 is another widely used dataset for evaluating entity disambiguation systems. In contrast with CoNLL, the mentions are sparsely annotated among documents. It contains 1074 annotated entities in the training set and 1020 in the evaluation set. The entities for this dataset are part of the TAC Knowledge Base, containing 818,741 entities. Due to the reduced entity vocabulary, we can fine-tune without resorting to an alias table and we adopt this setting throughout our results. This is consistent with the prior state-of-the-art approach of \citealt{zero_shot_el}. To select the context for a mention, we take the 256 bytes before and after the first occurrence of the mention in the document.

We select the fine-tuning parameters on training by doing cross-validation on the training set. We used a batch size of 32, trained for 1,000 steps and found it was best to freeze the entity embeddings. Our final model is trained on all the training data with the parameters selected in cross-validation. However, we report the result on the evaluation (test) set number in all tables, including ablations. Indeed, we found that this was more reflective of task performance, as the training set is significantly easier.

\subsection{End-to-end entity linking}

We also experiment with end-to-end entity linking on CoNLL (TAC-KBP is not suitable due to its sparse annotations). In this case, we do not use an alias table. In this setting, we follow the hyperparameters of Section~\ref{sec:all_vs_candidates}. Instead of using candidates, we train our model to predict \textsc{BIO}-tagged mention boundaries and to disambiguate among all entities. At training and fine-tuning time, gold spans are used for the disambiguation task.
At inference, we use the BIO-tagged predictions as our spans and predict entities for each span among all possible entities. We use the standard strong matching micro-F1 score.

\section{Evaluation}\label{sec:evaluation}

\subsection{Entity Linking}\label{sec:ent-linking-results}

Table~\ref{tab:entity_disambiguation} shows that our approach outperforms all prior approaches on CoNLL and TAC-KBP 2010. On CoNLL, we outperform methods in both alias-table settings. Additionally, we note that unlike many previous systems, we do not use
alias priors, knowledge-base features, or other entity features.

\begin{table}[]
    \centering
    \begin{tabular}{|l|c|c|c|c}
    \hline
System & CoNLL H & CoNLL P & TAC-KBP 2010 \\\hline
\citealt{chisholm2015entity} &  88.7 & - & 80.7 \\
\citealt{ganea2016probabilistic} & 87.6 & - & - \\
\citealt{globerson2016collective}& 91.0 & - & 87.2 \\
\citealt{pershina2015personalized}  & - & 91.8  & - \\
\citealt{globerson2016collective}   & - & 92.7  & 87.2 \\
\citealt{yamada2016joint}           & 91.5 & 93.1  & 85.2 \\
\citealt{raiman2018deeptype}$^{*}$        &  & 94.9  & 90.9  \\
\citealt{Yamada2017-dv}             & - & 94.3  & 87.7  \\
\citealt{ling2020learning}$\dagger$ & - & 94.9  & 89.8  \\
\citealt{zero_shot_el}              & - &  - & 94.0  \\
\hline
\ourapproach$\dagger$ & \textbf{92.5} & \textbf{96.7} & \textbf{94.9}\\
    \hline
    \end{tabular}
    \caption{Test accuracy on the CoNLL and TAC-KBP entity disambiguation tasks. CoNLL H refers to papers using the~\citealt{Hoffart2011-ip} ``means'' alias table while P refers to the~\citealt{pershina2015personalized} table. *It is not clear which alias table, if any, is used by \citealt{raiman2018deeptype}. $\dagger$-marked systems do not use features beyond the text and alias table.}
    \label{tab:entity_disambiguation}
\end{table}

\subsection{End-to-end entity linking}

For end-to-end entity linking, we do not use the alias table. Instead of using candidates, we predict BIO-tagged mention boundaries and disambiguate mentions among all entities. At training and fine-tuning time, gold spans are used for the disambiguation task. At inference, we use the BIO predictions as our spans and predict entities for all these spans.

Table~\ref{tab:e2e_linking} shows our model fares well against other models, with the exception of \citealt{broscheit} and \citealt{Kolitsas2018-dn}. The former use a much larger Transformer model, and also initialize from BERT-base model, which is pretrained on a corpus of unlabeled text much larger than our training data. \citealt{Kolitsas2018-dn} relies on an alias table to generate candidate mentions at both training and inference time. In addition, it introduces a clever mechanism to jointly optimize and select mention boundaries and entity candidates, whereas we use a simpler pipelined approach. Finally, they also introduce a document-level disambiguation coherence penalty and a coreference resolution heuristic. We believe the use of an alias table as well as the aforementioned differences explain the gap between our method and \citealt{Kolitsas2018-dn}, and we will look to bridge this gap in future work. Nevertheless, our model stands as a strong baseline of what can be achieved with simple modelling and low inference cost.

\begin{table}[]
    \centering
    \begin{tabular}{|l|c|c|}
    \hline
System                 & Development & Test    \\\hline
\citealt{Daiber2013-nr}   & 55.2   & 57.8   \\     
\citealt{Hoffart2011-ip}  & 72.4   & 72.8 \\
\citealt{tagme}           & 72.8   & 73.0 \\
\citealt{knowbert}        & 82.1   & 73.7   \\
\citealt{broscheit}       & 86.0   & 79.3   \\
\citealt{Kolitsas2018-dn} & 86.6   & 82.4   \\ \hline
\ourapproach              & 79.7   & 76.7     \\
    \hline
    \end{tabular}
    \caption{End-to-end entity linking strong matching micro-F1 score on the development and test sets of \conll. Despite the simplicity of our setup, our system is competitive with most prior work, with the noteworthy exception of \citealt{Kolitsas2018-dn}}.
    \label{tab:e2e_linking}
\end{table}

\section{Analysis}

\subsection{Classifying all entities or classifying candidates}\label{sec:all_vs_candidates}

We trained our model to distinguish the correct linked entity among candidates. An alternative approach is to predict among all entities. This is computationally more expensive as it requires doing a softmax over 5.7 million entities for every mention in the batch. Thus we use a batch size to 2048 and set the entity embedding dimension to 64 for both this model and the one trained with candidates. Table \ref{tab:all_vs_candidates} shows impressive accuracy without an alias table for the system classifying among all entities. However, it does not fare better than the model trained with candidates when using one. Given the considerable cost of doing the full sofmax for every mention, we use candidates for our other experiments. Note that our model gets 88.4\% accuracy when trained only on the CoNLL data.

\begin{table}[]
    \centering
    \begin{tabular}{|l|l|c|c|}
    \hline
Setup & No fine-tuning & Fine-tuned    \\\hline
Pre-trained and fine-tuned with all entities & 85.9 & 91.7  \\ \hline
No pre-training, fine-tuned with candidates & - & 88.4 \\
Pre-trained with all entities, fine-tuned with candidates & 92.4 & 97.2 \\
Pre-trained and fine-tuned with candidates & 92.6 & 97.1 \\
    \hline
    \end{tabular}
    \caption{Entity disambiguation accuracy on the \conll~development set for different pre-training and fine-tuning setups. Numbers in the first part of the table do not use an alias table, whereas the ones on the second part use~\citealt{pershina2015personalized}'s table.}
    \label{tab:all_vs_candidates}
\end{table}

\subsection{Impact of candidate selection}\label{sec:candidate_selection}

In Table~\ref{tab:candidate_selection}, we show the impact of different candidate selection pretraining strategies. On \conll, where we do use an alias table for evaluation, we find that our candidate selection heuristic seems to help the model in pre-training, achieving better performance than any of our ablations. Training with lexically related (phrase table) candidates is particularly important, as this is similar to the disambiguation task the model has to perform. Page candidates are semantically related but generally not lexically related and thus do not bring the same benefits. In fact, they might even distort the distribution of negative candidates as the model performs worse in this setting than with random candidates. After fine-tuning, all models fare similarly, which we believe is due to CoNLL having enough fine-tuning data so that all our models approach the performance upper-bound on this task (see Section~\ref{sec:qualitative_analysis}).  

For \tackbp, where we do not use an alias table at fine-tuning or inference time, the results are markedly different. After pre-training, \ourapproach~which has less random candidates performs worse than all other alternatives. This is likely because having more random candidates is closer to the full classification setup used in \tackbp. Given TAC-KBP's small training set, these differences carry over in fine-tuning performance.  

\begin{table}[]
    \centering
    \begin{tabular}{|l|c|c|c|c|}
    \hline
       & \multicolumn{2}{c|}{CoNLL} & \multicolumn{2}{c|}{TAC-KBP} \\ \hline
Candidates source & No fine-tuning & Fine-tuned & No fine-tuning & Fine-tuned  \\ \hline
\ourapproach & 92.2 & 96.9 & 87.3 & 91.4 \\
Phrase table and random & 88.0 & 96.9 & 91.6 & 94.7     \\
Page and random & 83.4 & 96.9 & 92.4 & 94.4   \\
Random & 85.3 & 97.2 & 91.7 & 94.9 \\
    \hline
    \end{tabular}
    \caption{Impact of the candidate selection method on development performance on CoNLL and \tackbp. Unsurprisingly, pretraining methods that are closer to the final task setup perform better: For CoNLL, we emphasize the importance of phrase table candidates in pre-training to emulate the use of the alias table , whereas for TAC-KBP, setups that use random candidates are more successful as they are closer to the full classification setup used in this task.}
    \label{tab:candidate_selection}
\end{table}

\subsection{Impact of adding noise during pretraining}\label{sec:noise}

Table~\ref{tab:noise} shows adding noise in pretraining helps both pre-training and fine-tuning performance. 
We hypothesize that input noise implicitly trains the model to generalize to alternative aliases: for instance, given the mention ``\texttt{Yuri Gagarin}'', the model might have to learn to recognize ``\texttt{Yuri [MASK]}'' or ``\texttt{[MASK] Gagarin}''.

\begin{table}[]
    \centering
    \begin{tabular}{|l|c|c|}
    \hline
System & No fine-tuning & Fine-tuned \\ \hline
With noise & 92.2 & 96.9 \\
Without noise & 88.1 & 96.1 \\
    \hline
    \end{tabular}
    \caption{Impact of adding BERT-style input noise during pre-training. We report development accuracy on CoNLL using the ~\citealt{pershina2015personalized} alias table.}
    \label{tab:noise}
\end{table}

Encouraged by the success of \citealt{devlin2018bert}, we experimented with also pretraining our model with a masked-language modelling objective, with the expectation that such pre-training would help our model learn better representations of language. We tried different architectures and loss weights (including a 4 layer transformer with both objectives at layer 4, a 12 layer with the entity linking loss at layer 4, etc.) but overall found this to not improve further on simply adding noise in pretraining.

\subsection{Impact of context selection methods on CoNLL}\label{sec:conll_context_selection}

Table~\ref{tab:context_ablation} shows the impact of varying CoNLL's context type in pre-training and fine-tuning. We find that larger contexts, especially those that include the beginning of the document, considerably boost performance before fine-tuning. However, similarly to our observations in Sections~\ref{sec:candidate_selection} and \ref{sec:noise}, we find that improvements are less marked after fine-tuning, likely because our performance is already very high.

\begin{table}[]
    \centering
       \begin{tabular}{|l|c|c|c|c}
       \hline
 & None & Title & Title \& first two sents\\ \hline
No Fine-tuning & 85.5 & 89.4 & 92.2  \\
Fine-tuned & 96.2 & 96.9  & 96.9  \\
       \hline
    \end{tabular}
    \caption{Impact of additional context beyond a single sentence used for entity disambiguation performance on the CoNLL task. We report development accuracy on CoNLL using the~\citealt{pershina2015personalized} alias table.}
    \label{tab:context_ablation}
\end{table}

\subsection{Error analysis}\label{sec:qualitative_analysis}

Figure~\ref{fig:conll_errors} shows three sample errors on the CoNLL development set. Most errors are due to varying levels of specificity in the CoNLL labels. Some errors are due to changes in Wikipedia. For instance, in text A, the Bulgaria U21 soccer team Wikipedia page was built in 2013, after CoNLL. 
Also, in text C, our model correctly disambiguates between Austin, Michaella and Richard Krajicek, which are all three tennis players (only Richard is a dutch).

\begin{figure*}[]
\begin{framed}
    \footnotesize
        \textbf{Text a}: Soccer - \textcolor{green}{Israel} beat \textcolor{red}{Bulgaria} in \textcolor{green}{European} under-21 qualifier. Herzliya, Israel 1996-08-31. \\
        \textcolor{red}{\textit{Incorrectly predicted}} ``Bulgaria'' as \texttt{Bulgaria national under-21 football team}. Second prediction is correct: \texttt{Bulgaria national football team}. \\
        \textcolor{green}{\textit{Correctly predicted}} ``Israel'' as \texttt{Israel national football team}. (There is no Wikipedia page for Israel national under-21 football team page). \textcolor{green}{\textit{Correctly predicted}} ``European''.\\
        \\ 
        \textbf{Text b}: Scottish \textcolor{red}{labour party} narrowly backs referendum. Stirling, Scotland 1996-08-31. Conservatives have only 10 of the 72 Scottish seats in parliament and consistently run third in opinion polls in Scotland behind labour and the independence-seeking Scottish national party.\\
        \textcolor{red}{\textit{Incorrectly predicted}} ``labour party'' as \texttt{Scottish Labour Party}. Gold Label: \texttt{Labour party (UK)}.\\
        \\ 
        \textbf{Text c}: \textcolor{green}{Edberg} refuses to qo (\emph{sic}) quietly. Richard Finn "it doesn't look all that bad" Edberg said of his path through the draw starting next with a match against \textcolor{green}{Krajicek}'s \textcolor{red}{dutch} countryman \textcolor{green}{Paul Haarhuis}.\\
        \textcolor{green}{\textit{Correctly predicted}} ``Edberg'' as \texttt{Stefan Edberg} and ``Krajicek'' as \texttt{Richard Krajicek} and ``Paul Haarhuis''. \\
        \textcolor{red}{\textit{Incorrectly predicted}} ``dutch'' as \texttt{Dutch people}. Second prediction is \texttt{Netherlands} and correct.
\end{framed} 
    \caption{Sample of errors on the CoNLL development set for our model.
    }
    \label{fig:conll_errors}
\end{figure*}

\section{Conclusion}

In this paper we present a thorough study of pretraining strategies for supervised entity linking, achieving state-of-the-art performance on both CoNLL and TAC-KBP 2010 with a four-layer Transformer-based model.
Given the limited headroom remaining in these datasets, and the strong impact of alias tables in simplifying the problem, we believe the creation of new datasets, and more difficult entity linking settings, such as zero-shot and low resource domains, are crucial areas for future work.

\acks{The authors wish to thank Dan Bikel and Eunsol Choi for their helpful comments in the preparation of this paper, as well as the anonymous reviewers.}

\newpage{}

\bibliography{eae}

\begin{thebibliography}{35}
\providecommand{\natexlab}[1]{#1}
\providecommand{\url}[1]{\texttt{#1}}
\expandafter\ifx\csname urlstyle\endcsname\relax
  \providecommand{\doi}[1]{doi: #1}\else
  \providecommand{\doi}{doi: \begingroup \urlstyle{rm}\Url}\fi

\bibitem[Broscheit(2019)]{broscheit}
Samuel Broscheit.
\newblock Investigating entity knowledge in bert with simple neural end-to-end
  entity linking.
\newblock \emph{CoNLL}, 2019.

\bibitem[Bunescu and Pasca(2006)]{bunescu2006using}
Razvan Bunescu and Marius Pasca.
\newblock Using encyclopedic knowledge for named entity disambiguation.
\newblock \emph{EACL}, 2006.

\bibitem[Cheng and Roth(2013)]{cheng2013relational}
Xiao Cheng and Dan Roth.
\newblock Relational inference for wikification.
\newblock In \emph{EMNLP}, 2013.

\bibitem[Chisholm and Hachey(2015)]{chisholm2015entity}
Andrew Chisholm and Ben Hachey.
\newblock Entity disambiguation with web links.
\newblock \emph{TACL}, 2015.

\bibitem[Daiber et~al.(2013)Daiber, Jakob, Hokamp, and Mendes]{Daiber2013-nr}
Joachim Daiber, Max Jakob, Chris Hokamp, and Pablo~N Mendes.
\newblock Improving efficiency and accuracy in multilingual entity extraction.
\newblock In \emph{ICSS}, 2013.

\bibitem[Devlin et~al.(2018)Devlin, Chang, Lee, and Toutanova]{devlin2018bert}
Jacob Devlin, Ming-Wei Chang, Kenton Lee, and Kristina Toutanova.
\newblock Bert: Pre-training of deep bidirectional transformers for language
  understanding.
\newblock \emph{arXiv preprint 1810.04805}, 2018.

\bibitem[F{\'e}vry et~al.(2020)F{\'e}vry, Soares, FitzGerald, Choi, and
  Kwiatkowski]{fevry2020eae}
Thibault F{\'e}vry, Livio~Baldini Soares, Nicholas FitzGerald, Eunsol Choi, and
  Tom Kwiatkowski.
\newblock Entities as experts: Sparse memory access with entity supervision.
\newblock \emph{arXiv preprint arXiv:2004.07202}, 2020.

\bibitem[Ganea et~al.(2016)Ganea, Ganea, Lucchi, Eickhoff, and
  Hofmann]{ganea2016probabilistic}
Octavian-Eugen Ganea, Marina Ganea, Aurelien Lucchi, Carsten Eickhoff, and
  Thomas Hofmann.
\newblock Probabilistic bag-of-hyperlinks model for entity linking.
\newblock In \emph{WWW}, 2016.

\bibitem[Gillick et~al.(2019)Gillick, Kulkarni, Lansing, Presta, Baldridge, Ie,
  and Garcia-Olano]{gillick2019learning}
Daniel Gillick, Sayali Kulkarni, Larry Lansing, Alessandro Presta, Jason
  Baldridge, Eugene Ie, and Diego Garcia-Olano.
\newblock Learning dense representations for entity retrieval.
\newblock \emph{arXiv preprint 1909.10506}, 2019.

\bibitem[Globerson et~al.(2016)Globerson, Lazic, Chakrabarti, Subramanya,
  Ringaard, and Pereira]{globerson2016collective}
Amir Globerson, Nevena Lazic, Soumen Chakrabarti, Amarnag Subramanya, Michael
  Ringaard, and Fernando Pereira.
\newblock Collective entity resolution with multi-focal attention.
\newblock \emph{ACL}, 2016.

\bibitem[Gupta et~al.(2017)Gupta, Singh, and Roth]{gupta2017entity}
Nitish Gupta, Sameer Singh, and Dan Roth.
\newblock Entity linking via joint encoding of types, descriptions, and
  context.
\newblock In \emph{EMNLP}, 2017.

\bibitem[Gutmann and Hyv{\"a}rinen(2012)]{gutmann2012noise}
Michael~U Gutmann and Aapo Hyv{\"a}rinen.
\newblock Noise-contrastive estimation of unnormalized statistical models, with
  applications to natural image statistics.
\newblock \emph{JMLR}, 2012.

\bibitem[Hoffart et~al.(2011)Hoffart, Yosef, Bordino, F{\"u}rstenau, Pinkal,
  Spaniol, Taneva, Thater, and Weikum]{Hoffart2011-ip}
Johannes Hoffart, Mohamed~Amir Yosef, Ilaria Bordino, Hagen F{\"u}rstenau,
  Manfred Pinkal, Marc Spaniol, Bilyana Taneva, Stefan Thater, and Gerhard
  Weikum.
\newblock Robust disambiguation of named entities in text.
\newblock In \emph{EMNLP}, 2011.

\bibitem[Ji et~al.(2010)Ji, Grishman, Dang, Griffitt, and
  Ellis]{ji2010overview}
Heng Ji, Ralph Grishman, Hoa~Trang Dang, Kira Griffitt, and Joe Ellis.
\newblock Overview of the tac 2010 knowledge base population track.
\newblock In \emph{TAC 2010}, 2010.

\bibitem[Kingma and Ba(2014)]{kingma2014adam}
Diederik~P Kingma and Jimmy Ba.
\newblock Adam: A method for stochastic optimization.
\newblock \emph{arXiv preprint 1412.6980}, 2014.

\bibitem[Kolitsas et~al.(2018)Kolitsas, Ganea, and Hofmann]{Kolitsas2018-dn}
Nikolaos Kolitsas, Octavian-Eugen Ganea, and Thomas Hofmann.
\newblock {End-to-End} neural entity linking.
\newblock In \emph{CoNLL}, 2018.

\bibitem[Lee et~al.(2016)Lee, Salant, Kwiatkowski, Parikh, Das, and
  Berant]{lee2016learning}
Kenton Lee, Shimi Salant, Tom Kwiatkowski, Ankur Parikh, Dipanjan Das, and
  Jonathan Berant.
\newblock Learning recurrent span representations for extractive question
  answering.
\newblock \emph{arXiv preprint 1611.01436}, 2016.

\bibitem[Ling et~al.(2020)Ling, FitzGerald, Shan, Soares, F{\'e}vry, Weiss, and
  Kwiatkowski]{ling2020learning}
Jeffrey Ling, Nicholas FitzGerald, Zifei Shan, Livio~Baldini Soares, Thibault
  F{\'e}vry, David Weiss, and Tom Kwiatkowski.
\newblock Learning cross-context entity representations from text.
\newblock \emph{arXiv preprint 2001.03765}, 2020.

\bibitem[Logeswaran et~al.(2019)Logeswaran, Chang, Lee, Toutanova, Devlin, and
  Lee]{logeswaran2019zero}
Lajanugen Logeswaran, Ming-Wei Chang, Kenton Lee, Kristina Toutanova, Jacob
  Devlin, and Honglak Lee.
\newblock Zero-shot entity linking by reading entity descriptions.
\newblock \emph{arXiv preprint 1906.07348}, 2019.

\bibitem[Luo et~al.(2015)Luo, Huang, Lin, and Nie]{luo2015joint}
Gang Luo, Xiaojiang Huang, Chin-Yew Lin, and Zaiqing Nie.
\newblock Joint entity recognition and disambiguation.
\newblock In \emph{EMNLP}, 2015.

\bibitem[Mihalcea and Csomai(2007)]{mihalcea2007wikify}
Rada Mihalcea and Andras Csomai.
\newblock Wikify! linking documents to encyclopedic knowledge.
\newblock In \emph{IKM}, 2007.

\bibitem[Mnih and Kavukcuoglu(2013)]{mnih2013learning}
Andriy Mnih and Koray Kavukcuoglu.
\newblock Learning word embeddings efficiently with noise-contrastive
  estimation.
\newblock In \emph{NeurIPS}, 2013.

\bibitem[Pershina et~al.(2015)Pershina, He, and
  Grishman]{pershina2015personalized}
Maria Pershina, Yifan He, and Ralph Grishman.
\newblock Personalized page rank for named entity disambiguation.
\newblock In \emph{NAACL}, 2015.

\bibitem[Peters et~al.(2019)Peters, Neumann, Logan, Robert, Schwartz, Joshi,
  Singh, and Smith]{knowbert}
Matthew~E Peters, Mark Neumann, IV~Logan, L~Robert, Roy Schwartz, Vidur Joshi,
  Sameer Singh, and Noah~A Smith.
\newblock Knowledge enhanced contextual word representations.
\newblock \emph{arXiv preprint 1909.04164}, 2019.

\bibitem[Piccinno and Ferragina(2014)]{tagme}
Francesco Piccinno and Paolo Ferragina.
\newblock From tagme to wat: a new entity annotator.
\newblock In \emph{Proceedings of the First International Workshop on Entity
  Recognition \& Disambiguation}, 2014.

\bibitem[Radhakrishnan et~al.(2018)Radhakrishnan, Talukdar, and
  Varma]{radhakrishnan}
Priya Radhakrishnan, Partha Talukdar, and Vasudeva Varma.
\newblock Elden: Improved entity linking using densified knowledge graphs.
\newblock In \emph{ACL-HLT}, 2018.

\bibitem[Raiman and Raiman(2018)]{raiman2018deeptype}
Jonathan~Raphael Raiman and Olivier~Michel Raiman.
\newblock Deeptype: multilingual entity linking by neural type system
  evolution.
\newblock In \emph{AAAI}, 2018.

\bibitem[Ringgaard et~al.(2017)Ringgaard, Gupta, and
  Pereira]{ringgaard2017sling}
Michael Ringgaard, Rahul Gupta, and Fernando~CN Pereira.
\newblock Sling: A framework for frame semantic parsing.
\newblock \emph{arXiv preprint 1710.07032}, 2017.

\bibitem[Sil and Yates(2013)]{sil2013re}
Avirup Sil and Alexander Yates.
\newblock Re-ranking for joint named-entity recognition and linking.
\newblock In \emph{IKM}, 2013.

\bibitem[Sil et~al.(2018)Sil, Kundu, Florian, and Hamza]{sil2018neural}
Avirup Sil, Gourab Kundu, Radu Florian, and Wael Hamza.
\newblock Neural cross-lingual entity linking.
\newblock In \emph{AAAI}, 2018.

\bibitem[Tsai and Roth(2016)]{tsai2016cross}
Chen-Tse Tsai and Dan Roth.
\newblock Cross-lingual wikification using multilingual embeddings.
\newblock In \emph{ACL-HLT}, 2016.

\bibitem[Vaswani et~al.(2017)Vaswani, Shazeer, Parmar, Uszkoreit, Jones, Gomez,
  Kaiser, and Polosukhin]{vaswani2017attention}
Ashish Vaswani, Noam Shazeer, Niki Parmar, Jakob Uszkoreit, Llion Jones,
  Aidan~N Gomez, {\L}ukasz Kaiser, and Illia Polosukhin.
\newblock Attention is all you need.
\newblock In \emph{NeurIPS}, 2017.

\bibitem[Wu et~al.(2019)Wu, Petroni, Josifoski, Riedel, and
  Zettlemoyer]{zero_shot_el}
Ledell Wu, Fabio Petroni, Martin Josifoski, Sebastian Riedel, and Luke
  Zettlemoyer.
\newblock Zero-shot entity linking with dense entity retrieval.
\newblock \emph{arXiv preprint 1911.03814}, 2019.

\bibitem[Yamada et~al.(2016)Yamada, Shindo, Takeda, and
  Takefuji]{yamada2016joint}
Ikuya Yamada, Hiroyuki Shindo, Hideaki Takeda, and Yoshiyasu Takefuji.
\newblock Joint learning of the embedding of words and entities for named
  entity disambiguation.
\newblock \emph{CoNLL}, 2016.

\bibitem[Yamada et~al.(2017)Yamada, Shindo, Takeda, and
  Takefuji]{Yamada2017-dv}
Ikuya Yamada, Hiroyuki Shindo, Hideaki Takeda, and Yoshiyasu Takefuji.
\newblock Learning distributed representations of texts and entities from
  knowledge base.
\newblock \emph{TACL}, 2017.

\end{thebibliography}
\bibliographystyle{plainnat}

\end{document}